\documentclass[11pt,a4paper]{article}
\usepackage{emnlp2018}

\usepackage{times}
\usepackage{latexsym}
\usepackage{multirow}
\usepackage{amsthm}
\usepackage{graphicx}
\usepackage{enumitem}
\usepackage{scalerel}
\usepackage{amsmath}
\usepackage{amssymb}
\usepackage{tkz-berge}

\usepackage{standalone}
\usepackage[normalem]{ulem}
\usepackage{color,soul}
\usepackage{arydshln}
\theoremstyle{theorem}
\newtheorem{theorem}{Theorem}[section]

\usepackage{url}

\aclfinalcopy 

    \setul{0.5ex}{0.3ex}
    \definecolor{Green}{rgb}{0,1,0}
    \setulcolor{Green}

    \setul{0.5ex}{0.3ex}
    \definecolor{Red}{rgb}{1,0.0,0.0}
    \setulcolor{Red}

    \setul{0.5ex}{0.3ex}
    \definecolor{Blue}{rgb}{0,0.0,1}
    \setulcolor{Blue}
    
\makeatletter
\def\squigglyred{\bgroup \markoverwith{\textcolor{red}{\lower3.5\p@\hbox{\sixly \char58}}}\ULon}
\def\squigglycyan{\bgroup \markoverwith{\textcolor{cyan}{\lower3.5\p@\hbox{\sixly \char58}}}\ULon}
\def\squigglybrown{\bgroup \markoverwith{\textcolor{brown}{\lower3.5\p@\hbox{\sixly \char58}}}\ULon}
\def\squigglypurple{\bgroup \markoverwith{\textcolor{purple}{\lower3.5\p@\hbox{\sixly \char58}}}\ULon}
\def\squigglyblue{\bgroup \markoverwith{\textcolor{blue}{\lower3.5\p@\hbox{\sixly \char58}}}\ULon}
\makeatother

\newcounter{Lcount}
\newcommand{\squishenum}{
\begin{list}{\arabic{Lcount}. }
{ \usecounter{Lcount}
\setlength{\itemsep}{0pt}
\setlength{\parsep}{0pt}
\setlength{\topsep}{0pt}
\setlength{\partopsep}{0pt}
\setlength{\leftmargin}{2em}
\setlength{\labelwidth}{1.5em}
\setlength{\labelsep}{0.5em} } }

\newcommand{\squishletter}{
\begin{list}{\alph{Lcount}. }
{ \usecounter{Lcount}
\setlength{\itemsep}{0pt}
\setlength{\parsep}{0pt}
\setlength{\topsep}{0pt}
\setlength{\partopsep}{0pt}
\setlength{\leftmargin}{2em}
\setlength{\labelwidth}{1.5em}
\setlength{\labelsep}{0.5em} } }

\newcommand{\squishlist}{
\begin{list}{$\bullet$}
{ \usecounter{Lcount}
\setlength{\itemsep}{0pt}
\setlength{\parsep}{0pt}
\setlength{\topsep}{0pt}
\setlength{\partopsep}{0pt}
\setlength{\leftmargin}{2em}
\setlength{\labelwidth}{1.5em}
\setlength{\labelsep}{0.5em} } }

\newcommand{\squishend}{
\end{list} }

\title{Neural Segmental Hypergraphs for Overlapping Mention Recognition}

\author{
Bailin Wang \\
University of Massachusetts \\ Amherst \\
  {\tt bailinwang@cs.umass.edu} 
  \And
Wei Lu  \\
Singapore University of Technology \\ and Design \\
  {\tt luwei@sutd.edu.sg} 
  }

\date{}

\begin{document}
\maketitle
\begin{abstract}
In this work, we propose a novel {\em segmental hypergraph} representation to model overlapping entity mentions that are prevalent in many practical datasets.
We show that our model built on top of such a new representation is able to capture features {\color{black}and interactions} that cannot be captured by previous models while maintaining a low time complexity for inference.
We also present a theoretical analysis to formally assess how our representation is better than alternative representations reported in the literature in terms of representational power.
Coupled with neural networks for feature learning, our model achieves the state-of-the-art performance in three benchmark datasets 
annotated with overlapping mentions.\footnote{{\color{black} We make our system and code available at: \url{http://statnlp.org/research/ie} }}
\end{abstract}

\section{Introduction}

One of the most crucial steps towards building a natural language understanding system is the identification of basic semantic chunks in text. 
Such a task is typically characterized by the named entity recognition task \cite{grishman1997information,tjong2003introduction}, or the more general mention recognition task, where mentions are defined as references to entities that could be named, nominal or pronominal \cite{florian2004statistical}.
The extracted mentions can be used in various downstream tasks for performing further semantic related tasks, including question answering \cite{abney2000answer}, relation extraction \cite{mintz2009distant,liu2017heterogeneous}, event extraction \cite{riedel2011fast,li-ji-huang:2013:ACL2013}, and coreference resolution \cite{soon2001machine,ng2002improving,chang2013constrained}.

One popular approach to the task of mention extraction is to regard it as a sequence labeling problem, with the underlying primary assumption being that the mentions are non-overlapping spans in the text.
However, as highlighted in several prior research efforts \cite{alex2007recognising,finkel2009nested,lu2015joint}, mentions may overlap with one another in practice.
Thus, models based on such a simplified assumption may result in sub-optimal performance for a down-stream task when they are deployed in practice.
For example, consider a phrase ``{\em At the Seattle zoo, \dots}'' shown in Figure \ref{fig:inst}, the relation \textsc{LocatedIn} between the  mentions ``{\em the Seattle zoo}'' (of type \textsc{Facility}) and ``{\em Seattle}'' (of type \textsc{Gpe}: Geo-political entities) will not be extracted unless both of these two overlapping mentions could be extracted.
Similarly, there are 4 mentions of the same type (\textsc{Protein}) in the text span ``{\em \dots\ PEBP2 alpha A1, alpha B1 \dots}'' taken from the biomedical domain. A downstream question answering system may fail to return the correct answer as desired, if the mention extraction system it relies on is unable to extract all these valid mentions.

\begin{figure}[t!]
\begin{center}
\scalebox{0.8}
{\footnotesize
\def\arraystretch{0.4}\tabcolsep=8pt
\setlength\doublerulesep{1pt}
\begin{tabular}{c}
\\
{\em \dots\ At $\underset{\text{\raisebox{2.5pt}[0pt][0pt]{\color{red}\scriptsize{}\textsc{Facility}}}}{\squigglyred{\text{the }\vphantom{\_}\underset{\text{\raisebox{2.5pt}[0pt][0pt]{\color{cyan}\scriptsize{}\textsc{Gpe}}}}{\squigglycyan{\text{Seattle}}}\text{ zoo}}}$}
{\em , efforts to  artificially  \dots}\\\\
\hline
\\
{\em \dots 
$
\underset{\text{\raisebox{2.5pt}[0pt][0pt]{\scriptsize\color{blue}\textsc{Protein}}}}
{\squigglyblue{\vphantom{\_}\underset{\text{\raisebox{2.5pt}[0pt][0pt]{\scriptsize\color{blue}\textsc{Protein}}}}
{\squigglyblue{\text{{\color{white}p}PEBP2}}}
\text{ }
\underset{\text{\raisebox{2.5pt}[0pt][0pt]{\scriptsize\color{blue}\textsc{Protein}}}}
{\squigglyblue{\text{alpha A1}}}
}}
$
}
{\em , 
$
{{\vphantom{\_}\underset{\text{\raisebox{2.5pt}[0pt][0pt]{\scriptsize\color{blue}\textsc{Protein}}}}{\squigglyblue{\text{alpha B1}}}\text{, and \dots }}}
$}\\
\end{tabular}
}\end{center}
\vspace{-3.5mm}
\caption{\color{black}Examples of overlapping mentions.}
\label{fig:inst}
\end{figure}

Various approaches to extracting overlapping mentions have been proposed in the past decade.
The cascaded approach \cite{alex2007recognising} builds a pipeline of sequence labeling models using conditional random fields (CRF) \cite{lafferty2001conditional}.
However, the model is unable to handle overlapping mentions of the same type.
\citet{finkel2009nested} presented a parsing based approach to nested mention extraction.
Due to the chart-based parsing algorithm involved, the model has a cubic time complexity in the number of words in the sentence.
A recent approach by \citet{lu2015joint} introduced a hypergraph representation for capturing overlapping mentions, which was shown fast and effective.
The work was improved by \citet{muis2017labeling}, who proposed a sequence labeling approach that assigns tags to gaps between words.
However, both approaches suffer from the {\em structural ambiguity} issue during inference, as we will further discuss in this paper.

We summarize our contributions as:

\squishenum

\item We propose a novel {\em segmental hypergraph} representation that is capable of modeling arbitrary combinations of (potentially overlapping) mentions in a given sentence. The model has a $\mathcal{O}(cmn)$ time complexity ($m$ is the number of mention types, $n$ is the number of words in a sentence, and $c$ is the maximal number of words for each mention), and is able to capture features that cannot be  captured by existing approaches.


\item Theoretically, we show that our approach based on such a new representation does not have the limitations associated with some recently proposed state-of-the-art approaches for overlapping mention extraction.

\item We show through extensive experiments on standard data that by exploiting both {\color{black}word-level} and span-level features learned from neural networks, our model is able to achieve the state-of-the-art performance for recognizing overlapping mentions.

\squishend

Our model is also general and robust. Further experiments show that our model yields competitive results when evaluated on data that does not have overlapping mentions annotated when comparing against other recently proposed state-of-the-art neural models that are capable of extracting non-overlapping mentions only.

\section{Related Work}

\subsection*{Overlapping Mention Recognition}

One of the earliest research efforts on handling overlapping mentions is a rule-based approach \cite{zhang2004enhancing,zhou2004recognizing,zhou2006recognizing} that is evaluated on the GENIA dataset \cite{kim2003genia}. 
The authors first detected the innermost mentions and then relied on rule-based post-processing methods to identify overlapping mentions. 
\citet{mcdonald2005flexible} presented a multilabel classification algorithm to model overlapping segments in a sentence systematically.

\citet{alex2007recognising} proposed several ways to combine multiple conditional random fields (CRF) \cite{lafferty2001conditional} for such tasks.
Their best results were obtained by cascading several CRF models in a specific order while each model is responsible for detecting mentions of a particular type.
Outputs of one model can also serve as features to the next model.
However, such an approach cannot model overlapping mentions of the same type, which frequently appear in practice.

\citet{finkel2009nested} approached this task from a parsing perspective by constructing a constituency tree, mapping each mention to a node in the tree.
This approach assumes one mention is {\em contained by} the other when they overlap.
While such an assumption largely holds in practice, it comes with a cost -- the chart-based parser suffers from its cubic time complexity, making it not scalable to large datasets involving long sentences.
Based on the same idea, \citet{nest-bw-em18} proposed a scalable transition-based approach to construct a constituency forest (a collection of constituency trees).

Instead of relying on structured models, \citet{xu2017local} proposed a local classifier for each possible span. 
However, this local approach is unable to capture the interactions between {\color{black}spans}.
Similar to \cite{alex2007recognising}, \citet{N18-1131} dynamically stacked multiple flat layers which recognize mentions sequentially from innermost mentions to outermost mentions.

Our work is inspired by the model of \citet{lu2015joint}, who introduced a {\em mention hypergraph} representation for capturing overlapping mentions.
Their model was shown fast and effective, and was improved by the mention separator model \cite{muis2017labeling}.
However, we note that (as also highlighted in their papers) both models suffer from the {\em structural ambiguity} issue during inference, which we will discuss later.
Our new  representation does not have this limitation.\footnote{A model comparison can be found later in Table \ref{tab:model_comparison}.}
Recently, \citet{N18-1079} also proposed a hypergraph-based representation based on the BILOU tagging scheme.
Their model is trained greedily using neural networks by viewing the hypergraph construction procedure as a multi-label assignment process.

\subsection*{Neural Models for Mention Recognition}

Recently, neural network based approaches to entity or mention recognition have received significant attention.
They have been proven effective, even in the absence of handcrafted features.
\citet{collobert2011natural} used convolutional neural networks (CNN) over word sequences, {\color{black} paired with a CRF output layer}.
\citet{huang2015bidirectional} replaced the CNN with a bidirectional long short-term memory network (LSTM) \cite{hochreiter1997long}.
\citet{strubell2017fast} proposed an iterated dilated CNN to improve computational efficiency.
Beyond word-level compositions, several methods incorporated character-level compositions with character embeddings, 
either through CNN \cite{chiu2016named,ma-hovy:2016:P16-1} or LSTM \cite{lample2016neural}.

\section{Segmental Hypergraph}

\begin{figure}[t!]
\small
\begin{center}
\scalebox{0.86}
{
\begin{tikzpicture}[scale=0.5, every node/.style={scale=.70}]

\tikzset{VertexStyle/.style={
  shape=circle,
  shading=ball,
  ball color=red!20,
  minimum size=18pt}
}

\Vertex[x=0.0,y=6.0,L=$\ \ \boldsymbol{\mathsf{A}}_i\ \ $]{A1}
\Vertex[x=3.0,y=6.0,L=$\boldsymbol{\mathsf{A}}_{i+1}$]{A2}
\Vertex[x=6.0,y=6.0,L=$\boldsymbol{\mathsf{A}}_{i+2}$]{A3}
\Vertex[x=9.0,y=6.0,L=$\boldsymbol{\mathsf{A}}_{i+3}$]{A4}

\tikzset{VertexStyle/.style={
  shape=circle,
  shading=ball,
  ball color=blue!20,
  minimum size=18pt}
}

\Vertex[x=0.0,y=4.0,L=$\ \ \boldsymbol{\mathsf{E}_i}\ \ $]{E1}
\Vertex[x=3.0,y=4.0,L=$\boldsymbol{\mathsf{E}_{i+1}}$]{E2}
\Vertex[x=6.0,y=4.0,L=$\boldsymbol{\mathsf{E}_{i+2}}$]{E3}
\Vertex[x=9.0,y=4.0,L=$\boldsymbol{\mathsf{E}_{i+3}}$]{E4}

\tikzset{VertexStyle/.style={
  shape=circle,
  shading=ball,
  ball color=green!20,
  minimum size=18pt}
}

\Vertex[x=1.0,y=2.5,L=$\ \boldsymbol{\mathsf{T}}^1_i\ $]{T1a}
\Vertex[x=1.0,y=-0.5,L=$\ \boldsymbol{\mathsf{T}}^k_i\ $]{T1b}
\Vertex[x=1.0,y=-3.5,L=$\ \boldsymbol{\mathsf{T}}^m_i\ $]{T1c}

\tikzset{VertexStyle/.style={
  shape=circle,
  shading=ball,
  ball color=gray!50,
  minimum size=18pt}
}

\Vertex[x=2.8,y=-0.5,L=$\boldsymbol{\mathsf{X}}$]{X}
\Vertex[x=2.5,y=-4.0,L=$\boldsymbol{\mathsf{X}}$]{X1}
\Vertex[x=5.5,y=-4.0,L=$\boldsymbol{\mathsf{X}}$]{X2}
\Vertex[x=8.5,y=-4.0,L=$\boldsymbol{\mathsf{X}}$]{X3}
\Vertex[x=11.5,y=-4.0,L=$\boldsymbol{\mathsf{X}}$]{X4}

\tikzset{VertexStyle/.style={
  shape=circle,
  shading=ball,
  ball color=yellow!20,
  minimum size=18pt}
}

\Vertex[x=2.5,y=-2.0,L=$\ \ \boldsymbol{\mathsf{I}}^k_{i,i}\ \ $]{I1b}
\Vertex[x=5.5,y=-2.0,L=$\boldsymbol{\mathsf{I}}^k_{i,i+1}$]{I2b}
\Vertex[x=8.5,y=-2.0,L=$\boldsymbol{\mathsf{I}}^k_{i,i+2}$]{I3b}
\Vertex[x=11.5,y=-2.0,L=$\boldsymbol{\mathsf{I}}^k_{i,i+3}$]{I4b}

\tikzset{VertexStyle/.style = {line width=1pt,scale=1.2}}
\Vertex[x=0.0,y=7.5,L=$i$]{Time1}
\Vertex[x=3.0,y=7.5,L=$i+1$]{Time2}
\Vertex[x=6.0,y=7.5,L=$i+2$]{Time3}
\Vertex[x=9.0,y=7.5,L=$i+3$]{Time4}
\Vertex[x=1.0,y=1.0,L=$\vdots$]{T1x}
\Vertex[x=1.0,y=-2.0,L=$\vdots$]{T1y}
\Vertex[x=4.0,y=2.5,L=$\ddots$]{T2x}
\Vertex[x=7.0,y=2.5,L=$\ddots$]{T3x}
\Vertex[x=10.0,y=2.5,L=$\ddots$]{T4x}
{\color{white}\Vertex[x=12.0,y=6.0,L=$\boldsymbol{\mathsf{A}}_{i+4}$]{A5}}

\draw[red][->]    (A1) to[out=-30,in=-160, draw=blue]  (A2);
\draw[red][->]    (A1) to[out=-30,in=60] (E1);

\draw[red][->]    (A2) to[out=-30,in=-160, draw=blue]  (A3);
\draw[red][->]    (A2) to[out=-30,in=60] (E2);

\draw[red][->]    (A3) to[out=-30,in=-160, draw=blue]  (A4);
\draw[red][->]    (A3) to[out=-30,in=60] (E3);

\draw[red,dashed][->]    (A4) to[out=-30,in=-160, draw=blue]  (A5);
\draw[red][->]    (A4) to[out=-30,in=60] (E4);

\draw[blue][->]    (E1) to[out=-90,in=180] (T1a);
\draw[blue][->]   (E1) to[out=-90,in=150] (T1x);
\draw[blue][->]    (E1) to[out=-90,in=140] (T1b);
\draw[blue][->]    (E1) to[out=-90,in=170] (T1y);
\draw[blue][->]    (E1) to[out=-90,in=180] (T1c);

\draw[orange][->]     (T1b) to[out=0,in=180] (X);

\draw[blue!50!red][->]     (T1b) to[out=-60,in=150] (I1b);

\draw[green][->]    (I1b) to[out=0,in=180] (I2b);
\draw[green][->]    (I2b) to[out=0,in=180] (I3b);
\draw[green][->]    (I3b) to[out=0,in=180] (I4b);

\draw[black][->]    (I1b) to[out=-90,in=90] (X1);
\draw[black][->]    (I2b) to[out=-90,in=90] (X2);
\draw[black][->]    (I3b) to[out=-90,in=90] (X3);
\draw[black][->]    (I4b) to[out=-90,in=90] (X4);

\draw[green!40!red][->]    (I1b) to[out=-30,in=-150] (I2b);
\draw[green!40!red][->]    (I2b) to[out=-30,in=-150] (I3b);
\draw[green!40!red][->]    (I3b) to[out=-30,in=-150] (I4b);

\draw[green!40!red][->]   (I1b) to[out=-30,in=30] (X1);
\draw[green!40!red][->]   (I2b) to[out=-30,in=30] (X2);
\draw[green!40!red][->]   (I3b) to[out=-30,in=30] (X3);

\end{tikzpicture}
}
\caption{
An example of partial segmental hypergraph (hyperedges of different types in different colors).
}
\label{fig:arch}
\end{center}
\end{figure}
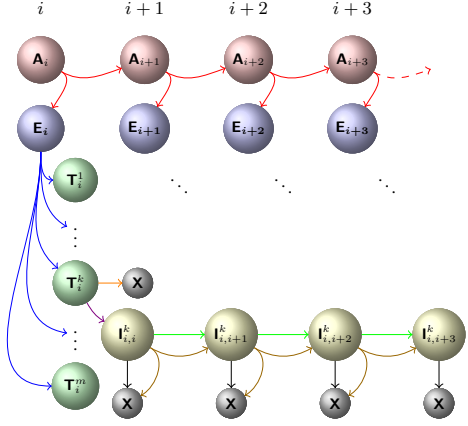


A segmental hypergraph is a representation that aims at representing all possible combinations of (potentially overlapping) mentions in a given sentence.
It belongs to a  class of directed hypergraphs \cite{gallo1993directed}, where each hyperedge $e$ consists of a single designated parent node ({\em head} of $e$) and an ordered list of child nodes ({\em tail} of $e$). 
Specifically, our segmental hypergraph consists of the following 5 types of nodes:
\squishlist
\item $\boldsymbol{\mathsf{A}}_i$ encodes all such mentions that start with the $i$-th or a later word
\item $\boldsymbol{\mathsf{E}}_i$ encodes all  mentions that start exactly with the $i$-th word
\item $\boldsymbol{\mathsf{T}}^k_i$ represents all mentions of type $k$ starting with the $i$-{th} word
\item $\boldsymbol{\mathsf{I}}^k_{i,j}$ represents all mentions of type $k$ that contain the $j$-th word and start with the $i$-th word
\item $\boldsymbol{\mathsf{X}}$ marks the end of a mention.
\squishend

Hyperedges connecting these nodes are designed to indicate how the semantics of a parent node can be re-expressed in terms of its child nodes.
Figure \ref{fig:arch} gives a partial segmental hypergraph representing all combinations of mentions within the  span $[i, i+3]$ consisting of 4 {\color{black}words}.
There are 4  types of hyperedges:
\squishenum
\item  A hyperedge $\{\boldsymbol{\mathsf{A}}_i \rightarrow (\boldsymbol{\mathsf{A}}_{i+1},\boldsymbol{\mathsf{E}}_i)\}$ from $\boldsymbol{\mathsf{A}}_i$ to its children implies the fact that $\boldsymbol{\mathsf{A}}_i$ consists of those mentions that either ``start exactly with the $i$-th word" ($\boldsymbol{\mathsf{E}}_i$), or ``start with a word that appears strictly after the $i$-th word" ($\boldsymbol{\mathsf{A}}_{i+1}$).
\item  A hyperedge $\{\boldsymbol{\mathsf{E}}_i \rightarrow (\boldsymbol{\mathsf{T}}^1_i,\dots,\boldsymbol{\mathsf{T}}^m_i)\}$ from $\boldsymbol{\mathsf{E}}_i$ to its children implies that we should consider all possible types for the mentions (possibly of length 0) that start with the $i$-th word.
\item Two hyperedges $\{\boldsymbol{\mathsf{T}}^k_i \rightarrow \boldsymbol{\mathsf{I}}^k_{i,i}\}$ and $\{\boldsymbol{\mathsf{T}}^k_i \rightarrow \boldsymbol{\mathsf{X}}\}$ from $\boldsymbol{\mathsf{T}}^k_i$ indicate that either there  exists at least one mention starting  with the $i$-th word (the former hyperedge), or there does not exist any such mention (the latter hyperedge).
\item  Three hyperedges $\{\boldsymbol{\mathsf{I}}^k_{i,j} \rightarrow \boldsymbol{\mathsf{I}}^k_{i,j+1}\}$, $\{\boldsymbol{\mathsf{I}}^k_{i,j} \rightarrow \boldsymbol{\mathsf{X}}\}$, and $\{\boldsymbol{\mathsf{I}}^k_{i,j} \rightarrow (\boldsymbol{\mathsf{I}}^k_{i,j+1}, \boldsymbol{\mathsf{X}})\}$ from $\boldsymbol{\mathsf{I}}^k_{i,j}$ indicate the following three cases respectively: 1) both the $j$-th and $(j+1)$-th words belong to at least one mention that starts with the $i$-th word, 2) there exists one mention that starts with the $i$-th word and ends with the $j$-th word, and 3) both cases are valid.
\squishend

Essentially, the complete hypergraph compactly encodes the whole search space of all possible mentions that can ever appear within a sentence, where such mentions may or may not overlap with one another.
When we traverse the complete segmental hypergraph by following the directions as specified by the hyperedges, selecting only one outgoing hyperedge at a time at each node, we arrive at a {\em hyperpath}\footnote{Each hyperpath is a {\em hypertree} \cite{brandstadt1998dually}.} -- a rooted, directed sub-structure contained by the original hypergraph.

\begin{figure}[t!]
\begin{center}
\scalebox{0.7}
{
\begin{tikzpicture}[scale=0.5, every node/.style={scale=.70}]

\tikzset{VertexStyle/.style={
  shape=circle,
  shading=ball,
  ball color=red!20,
  minimum size=18pt}
}

\Vertex[x=0.0,y=10.0,L=$\boldsymbol{\mathsf{A}}_{1}$]{A1}
\Vertex[x=4.0,y=10.0,L=$\boldsymbol{\mathsf{A}}_{2}$]{A2}
\Vertex[x=8.0,y=10.0,L=$\boldsymbol{\mathsf{A}}_{3}$]{A3}
\Vertex[x=12.0,y=10.0,L=$\boldsymbol{\mathsf{A}}_{4}$]{A4}

\tikzset{VertexStyle/.style={
  shape=circle,
  shading=ball,
  ball color=blue!20,
  minimum size=18pt}
}

\Vertex[x=0.0,y=8.0,L=$\boldsymbol{\mathsf{E}_{1}}$]{E1}
\Vertex[x=4.0,y=8.0,L=$\boldsymbol{\mathsf{E}_{2}}$]{E2}
\Vertex[x=8.0,y=8.0,L=$\boldsymbol{\mathsf{E}_{3}}$]{E3}
\Vertex[x=12.0,y=8.0,L=$\boldsymbol{\mathsf{E}_{4}}$]{E4}

\tikzset{VertexStyle/.style={
  shape=circle,
  shading=ball,
  ball color=green!20,
  minimum size=18pt}
}

\Vertex[x=0.0,y=6,L=$\ \boldsymbol{\mathsf{T}}^1_1\ $]{T1a}
\Vertex[x=0.0,y=1,L=$\ \boldsymbol{\mathsf{T}}^2_1\ $]{T1b}
\Vertex[x=4,y=6,L=$\ \boldsymbol{\mathsf{T}}^1_2\ $]{T2a}
\Vertex[x=4,y=1,L=$\ \boldsymbol{\mathsf{T}}^2_2\ $]{T2b}
\Vertex[x=8,y=6,L=$\ \boldsymbol{\mathsf{T}}^1_3\ $]{T3a}
\Vertex[x=8,y=1,L=$\ \boldsymbol{\mathsf{T}}^2_3\ $]{T3b}
\Vertex[x=12,y=6,L=$\ \boldsymbol{\mathsf{T}}^1_4\ $]{T4a}
\Vertex[x=12,y=1,L=$\ \boldsymbol{\mathsf{T}}^2_4\ $]{T4b}

\tikzset{VertexStyle/.style={
  shape=circle,
  shading=ball,
  ball color=gray!50,
  minimum size=18pt}
}

\Vertex[x=1.6,y=6,L=$\boldsymbol{\mathsf{X}}$]{XT1a}
\Vertex[x=5.6,y=1,L=$\boldsymbol{\mathsf{X}}$]{XT2b}
\Vertex[x=9.6,y=6,L=$\boldsymbol{\mathsf{X}}$]{XT3a}
\Vertex[x=9.6,y=1,L=$\boldsymbol{\mathsf{X}}$]{XT3b}
\Vertex[x=13.6,y=6,L=$\boldsymbol{\mathsf{X}}$]{XT4a}
\Vertex[x=13.6,y=1,L=$\boldsymbol{\mathsf{X}}$]{XT4b}

\Vertex[x=8,y=-3.0,L=$\boldsymbol{\mathsf{X}}$]{X13}
\Vertex[x=12,y=-3.0,L=$\boldsymbol{\mathsf{X}}$]{X14}
\Vertex[x=5.6,y=4.0,L=$\boldsymbol{\mathsf{X}}$]{X22}


\tikzset{VertexStyle/.style={
  shape=circle,
  shading=ball,
  ball color=yellow!20,
  minimum size=18pt}
}

\Vertex[x=0,y=-1.0,L=$\boldsymbol{\mathsf{I}}^2_{1,1}$]{I11b}
\Vertex[x=4,y=-1,L=$\boldsymbol{\mathsf{I}}^2_{1,2}$]{I12b}
\Vertex[x=8,y=-1.0,L=$\boldsymbol{\mathsf{I}}^2_{1,3}$]{I13b}
\Vertex[x=12,y=-1.0,L=$\boldsymbol{\mathsf{I}}^2_{1,4}$]{I14b}

\Vertex[x=4,y=4,L=$\boldsymbol{\mathsf{I}}^1_{2,2}$]{I22a}

\tikzset{VertexStyle/.style = {line width=1pt,scale=1.2}}
\Vertex[x=0.0,y=11.2,L={\em Israeli}]{Time1}
\Vertex[x=4.0,y=11.2,L={\em UN}]{Time2}
\Vertex[x=8.0,y=11.2,L={\em Ambassador}]{Time3}
\Vertex[x=12.0,y=11.2,L={\em Danny}]{Time4}

\Vertex[x=0,y=-4,L=\small{[}]{}
\Vertex[x=8,y=-4,L=\small{]}]{}
\Vertex[x=4,y=-4,L=\small{\textsc{Person}}]{}

\Vertex[x=0,y=-4.8,L=\small{[}]{}
\Vertex[x=12,y=-4.8,L=\small{]}]{}
\Vertex[x=5.5,y=-4.8,L=\small{\textsc{Person}}]{}

\Vertex[x=3.2,y=2.8,L=\small{[}]{}
\Vertex[x=6.8,y=2.8,L=\small{]}]{}
\Vertex[x=5,y=2.8,L=\small{\textsc{Organization}}]{}

\draw[red][->]    (A1) to[out=-30,in=-160, draw=blue]  (A2);
\draw[red][->]    (A1) to[out=-30,in=60] (E1);

\draw[red][->]    (A2) to[out=-30,in=-160, draw=blue]  (A3);
\draw[red][->]    (A2) to[out=-30,in=60] (E2);

\draw[red][->]    (A3) to[out=-30,in=-160, draw=blue]  (A4);
\draw[red][->]    (A3) to[out=-30,in=60] (E3);

\draw[red][->]    (A4) to[out=-30,in=60] (E4);

\draw[blue][->]    (E1) to[out=-145,in=140] (T1a);
\draw[blue][->]    (E1) to[out=-145,in=120] (T1b);
\draw[blue][->]    (E2) to[out=-145,in=140] (T2a);
\draw[blue][->]    (E2) to[out=-145,in=120] (T2b);
\draw[blue][->]    (E3) to[out=-145,in=140] (T3a);
\draw[blue][->]    (E3) to[out=-145,in=120] (T3b);
\draw[blue][->]    (E4) to[out=-145,in=140] (T4a);
\draw[blue][->]    (E4) to[out=-145,in=120] (T4b);

\draw[orange][->]    (T1a) to[out=0,in=180] (XT1a);
\draw[orange][->]    (T2b) to[out=0,in=180] (XT2b);
\draw[orange][->]    (T3a) to[out=0,in=180] (XT3a);
\draw[orange][->]    (T3b) to[out=0,in=180] (XT3b);
\draw[orange][->]    (T4a) to[out=0,in=180] (XT4a);
\draw[orange][->]    (T4b) to[out=0,in=180] (XT4b);

\draw[blue!50!red][->]    (T1b) to[out=-90,in=90] (I11b);
\draw[blue!50!red][->]    (T2a) to[out=-90,in=90] (I22a);

\draw[green][->]     (I11b) to[out=0,in=180] (I12b);
\draw[green][->]     (I12b) to[out=0,in=180] (I13b);
\draw[black][->]     (I14b) to[out=-90,in=90] (X14);
\draw[green!40!red][->]     (I13b) to[out=-30,in=-150] (I14b);
\draw[green!40!red][->]    (I13b) to[out=-30,in=30] (X13);

\draw[black][->]    (I22a) to[out=0,in=-180] (X22);

\end{tikzpicture}
}
\caption{
A specific  hyperpath for encoding three mentions.
For brevity, we only show two types.
}
\label{fig:example}
\end{center}
\end{figure}
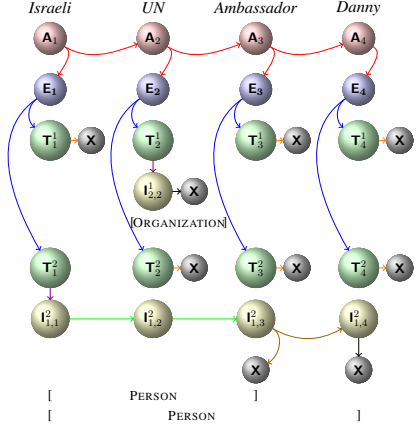

Figure \ref{fig:example} shows an example. 
Here, ``\textit{Israeli UN Ambassador}" of type \textsc{Person} is captured by the following sequence of nodes (along a hyperpath):
 ``$\boldsymbol{\mathsf{A}}_1$, $\boldsymbol{\mathsf{E}}_1$, $\boldsymbol{\mathsf{T}}^2_1$, $\boldsymbol{\mathsf{I}}^2_{1,1}$, $\boldsymbol{\mathsf{I}}^2_{1,2}$, $\boldsymbol{\mathsf{I}}^2_{1,3}$, $\boldsymbol{\mathsf{X}}$'', while 
 ``\textit{Israeli UN Ambassador Danny}" of type \textsc{Person} corresponds to the following node sequence: ``$\boldsymbol{\mathsf{A}}_1$, $\boldsymbol{\mathsf{E}}_1$, $\boldsymbol{\mathsf{T}}^2_1$, $\boldsymbol{\mathsf{I}}^2_{1,1}$, $\boldsymbol{\mathsf{I}}^2_{1,2}$, $\boldsymbol{\mathsf{I}}^2_{1,3}$, $\boldsymbol{\mathsf{I}}^2_{1,4}$, $\boldsymbol{\mathsf{X}}$''.
Similarly, the following sequence ``$\boldsymbol{\mathsf{A}}_1$, $\boldsymbol{\mathsf{A}}_2$, $\boldsymbol{\mathsf{E}}_2$, $\boldsymbol{\mathsf{T}}^1_2$, $\boldsymbol{\mathsf{I}}^1_{2,2}$, $\boldsymbol{\mathsf{X}}$'' represents the mention ``\textit{UN}'' of type \textsc{Organization}.
As we can see, such node sequences together form a single hyperpath that encodes this specific combination of mentions that overlap with one another.
More details on segmental hypergraph and hyperpaths are in the supplementary material.

\begin{table*}[h!]
\begin{center}
\scalebox{0.7}
{
\begin{tabular}{rp{20mm}p{20mm}p{20mm}p{20mm}p{20mm}p{20mm}}

&&&&&&\\
\multicolumn{1}{c|}{}&\multicolumn{1}{c|}{Spurious}&\multicolumn{1}{c|}{Structural} & \multicolumn{1}{c|}{Only Nested} &\multicolumn{1}{c|}{Pipeline}&\multicolumn{1}{c|}{Different} &\multicolumn{1}{c}{Time}
\\
\multicolumn{1}{c|}{}&\multicolumn{1}{c|}{Structures}&\multicolumn{1}{c|}{Ambiguity}& \multicolumn{1}{c|}{Mentions} &\multicolumn{1}{c|}{Approach} &\multicolumn{1}{c|}{Types Only}& \multicolumn{1}{c}{Complexity}
\\
\hline
{Alex et al. \citeyearpar{alex2007recognising}}
&\multicolumn{1}{|c|}{\textsc{No}}&\multicolumn{1}{c|}{\textsc{No}}&\multicolumn{1}{c|}{\textsc{No}}&\multicolumn{1}{c|}{\textsc{Yes}}&\multicolumn{1}{c|}{\textsc{Yes}}&\multicolumn{1}{c}{$\mathcal{O}(mn)$}
\\
{Finkel \& Manning \citeyearpar{finkel2009nested}}
&\multicolumn{1}{|c|}{\textsc{No}}&\multicolumn{1}{c|}{\textsc{No}}&\multicolumn{1}{c|}{\textsc{Yes}}&\multicolumn{1}{c|}{\textsc{No}}&\multicolumn{1}{c|}{\textsc{No}}&\multicolumn{1}{c}{$\mathcal{O}(|G|n^3)$}
\\
{Lu \& Roth \citeyearpar{{lu2015joint}}}
&\multicolumn{1}{|c|}{\textsc{Yes}}&\multicolumn{1}{c|}{\textsc{Yes}}&\multicolumn{1}{c|}{\textsc{No}}&\multicolumn{1}{c|}{\textsc{No}}&\multicolumn{1}{c|}{\textsc{No}}&\multicolumn{1}{c}{$\mathcal{O}(mn)$}
\\
{Muis \& Lu \citeyearpar{muis2017labeling}}
&\multicolumn{1}{|c|}{\textsc{No}}&\multicolumn{1}{c|}{\textsc{Yes}}&\multicolumn{1}{c|}{\textsc{No}}&\multicolumn{1}{c|}{\textsc{No}}&\multicolumn{1}{c|}{\textsc{No}}&\multicolumn{1}{c}{$\mathcal{O}(mn)$}
\\
{Wang et al. \citeyearpar{nest-bw-em18}}
&\multicolumn{1}{|c|}{\textsc{No}}&\multicolumn{1}{c|}{\textsc{No}}&\multicolumn{1}{c|}{\textsc{Yes}}&\multicolumn{1}{c|}{\textsc{No}}&\multicolumn{1}{c|}{\textsc{No}}&\multicolumn{1}{c}{$\mathcal{O}(mn)$}
\\
{This work}
&\multicolumn{1}{|c|}{\bf \textsc{No}}&\multicolumn{1}{c|}{\bf \textsc{No}}&\multicolumn{1}{c|}{\bf \textsc{No}}&\multicolumn{1}{c|}{\bf \textsc{No}}&\multicolumn{1}{c|}{\bf \textsc{No}}&\multicolumn{1}{c}{\bf $\mathcal{O}(cmn)$}

\\
\end{tabular}
}
\end{center}
\vspace{-4mm}
\caption{Model comparison. $|G|$ is the number of  rules in grammar $G$.
}
\label{tab:model_comparison}
\end{table*}

\subsection*{Theoretical Analysis}

Our proposed segmental hypergraph representation has the following theoretical property:
\theoremstyle{theorem}
\begin{theorem}
{\bf (Structural Ambiguity Free)} 

For any sentence and its segmental hypergraph $\mathcal{G}=(\mathcal{V}, \mathcal{E})$, let $\mathcal{S}$ be the set of all possible mention combinations for the given sentence, and $\mathcal{P}$ be the set of all hyperpaths contained by $\mathcal{G}$, there is a one-to-one correspondence between elements in $\mathcal{P}$ and $\mathcal{S}$.

\end{theorem}

Due to space, we provide a proof sketch and include more details in the supplementary material.

\noindent \textbf{Proof Sketch}
{\color{black}
We note that each hyperpath is uniquely characterized by its collection of hyperedges that involve $\boldsymbol{\mathsf{X}}$ nodes. These hyperedges uniquely determine the collection of mentions. Conversely, a collection of mentions can be uniquely characterized by a collection of such hyperedges, which yields a unique hyperpath.
}
$\qed$

Note that such a theorem states that our novel representation has no {\em structural ambiguity}, 
a nice property that both mention hypergraph model of \cite{lu2015joint} and mention separator model of \cite{muis2017labeling} do not hold. 
As the authors have mentioned in their papers, for a given sub-structure in their model, there exist multiple ways of interpreting the combination of mentions.
Specifically, in both representations, the decisions on where the {\em beginning} and the {\em end} of a mention are made {\em locally}.
Such a design will lead to the structural ambiguity as there will be multiple interpretations to the mentions given a particular collection of positions marked as beginning and end of mentions.
To illustrate, consider a phrase with 4 words ``{\em A B C D}'' where there are only two overlapping mentions ``{\em B C}'' and ``{\em A B C D}''.
In both of the previous approaches, their models would make local predictions and assign both ``{\em A}'' and ``{\em B}'' as left boundaries, and both ``{\em C}'' and ``{\em D}'' as right boundaries. 
However, based on such local predictions one could also interpret ``{\em A B C}'' as a mention -- this is where the ambiguity arises.
In contrast, our model enjoys the structural ambiguity free property as it uses our newly defined $\boldsymbol{\mathsf{I}}$ nodes (together with $\boldsymbol{\mathsf{X}}$ nodes) to jointly capture the complete boundary information of mentions.

Table \ref{tab:model_comparison} shows a full comparison.
\footnote{The mention hypergraph \cite{lu2015joint} also suffers from the {\em spurious structures} issue, while we do not. We refer the readers to \cite{muis2017labeling} for details.}

\section{Learning}



We adopt a log-linear approach to model the conditional probability of each hyperpath as follows:
\begin{center}
\vspace{-3mm}
\small
\begin{equation}
p(\boldsymbol{y} |\boldsymbol{x}) = \frac{\exp  f(\boldsymbol{x},\boldsymbol{y})}{\sum_{\boldsymbol{y}'} \exp f(\boldsymbol{x},\boldsymbol{y}')}
\label{eq:loglinear}
\end{equation}
\vspace{-3mm}
\end{center}
where $f(\boldsymbol{x},\boldsymbol{y})$ is the score function for any pair of input sentence $\boldsymbol{x}$ and output mention combination $\boldsymbol{y}$, which corresponds to a unique hyperpath $\mathcal{G}_{\boldsymbol{y}}$. 
Our objective is to minimize the negative log-likelihood of all instances in the training set $\mathcal{D}$:
\begin{center}
\vspace{-3mm}
\small
\begin{equation}
-\sum_{(\boldsymbol{x}, \boldsymbol{y}^*)\in\mathcal{D}} \log p(\boldsymbol{y}^*|\boldsymbol{x})
\end{equation}
\vspace{-3mm}
\end{center}

We define features over each hyperedge within the hyperpath $\mathcal{G}_{\boldsymbol{y}}$.
The score function can be decomposed into the following form:
\begin{center}
\vspace{-3mm}
\small
\begin{align}
f(\boldsymbol{x},\boldsymbol{y})
=&
\sum_{e\in \mathcal{G}_{\boldsymbol{y}}} \psi(e, \boldsymbol{x})
\end{align}
\vspace{-3mm}
\end{center}
where $e\in \mathcal{G}_{\boldsymbol{y}}$ denotes a hyperedge that appears within the hyperpath $\mathcal{G}_{\boldsymbol{y}}$, and $\psi(e,\boldsymbol{x})$ is a score defined over   $e$ when the input sentence is $\boldsymbol{x}$.

{\color{black}
Apart from {\em word-level} features, the segmental hypergraph also allows {\em span-level} features to be defined.
The node ${\boldsymbol{\mathsf{I}}^k_{i, j}}$ corresponds to a particular span $[i, j]$ over which we can extract our local features.
The hyperedge between ${\boldsymbol{\mathsf{I}}}$ nodes can capture the interactions between partial mentions and 
hyperedge between ${\boldsymbol{\mathsf{I}}^k_{i, j}}$ and $\boldsymbol{\mathsf{X}}$ precisely represents the mention $[i, j]$ with type $k$.
We note that such features and interactions cannot be captured by the models of \cite{lu2015joint} and \cite{muis2017labeling}.
Such a unique property makes our segmental hypergraph model more expressive than theirs.
}

\subsection{Softmax-Margin Training}

Inspired by \cite{mohit2012recall}, we consider the softmax-margin \cite{gimpel-smith:2010:NAACLHLT} in our model.
The function $\psi(e,\boldsymbol{x})$ is defined as follows:
\begin{center}
\vspace{-3mm}
\small
\begin{align}
\psi(e, \boldsymbol{x})
=
\phi(e, \boldsymbol{x}) + \Delta(e, \mathcal{G}_{\boldsymbol{y^*}})
\end{align}
\end{center}
where $\phi(e, \boldsymbol{x})$ is a feature function, and $\Delta(e,\mathcal{G}_{\boldsymbol{y}^*})$ is the cost function that defines the margin:
\begin{center}
\vspace{-3mm}
\small
\begin{align}
\Delta(e, \mathcal{G}_{\boldsymbol{y^*}} ) =  {\left\{ \begin{array}{l}
\beta \ \quad \mathrm{TX}[e] \wedge e \notin \mathcal{G}_{\boldsymbol{y^*}} \\
 1 \quad \mathrm{TI}[e] \wedge e \notin \mathcal{G}_{\boldsymbol{y^*}} \\  
 0 \quad  \quad \quad \textnormal{otherwise}
\end{array} \right.} 
\end{align}
\vspace{-3mm}
\end{center}

Here, $\boldsymbol{y}^*$ is the gold mention combination, and $\mathrm{TX}[e]$ and $\mathrm{TI}[e]$ are indicator functions that  return true if $e$ is between $\boldsymbol{\mathsf{T}$ and $\mathsf{X}}$ and between $\boldsymbol{\mathsf{T}$ and $\mathsf{I}}$ respectively, and false otherwise. 
We set $\beta\geq 1$ such that the cost function will assign more penalty to false negatives than to false positives.

\subsection{Feature Representation}


We use two bidirectional LSTMs to learn word-level and span-level feature representations that can be used in our approach, resulting in our {\em neural segmental hypergraph} model.
We first map the $i$-th word in a sentence to its pre-trained word embedding $\mathbf{e}_i$, and its POS tag to its embedding $\mathbf{p}_i$ if it exists.
The final representation for $i$-{th} word is the concatenation of them: $\mathbf{v}_i= [\mathbf{e}_i, \mathbf{p}_i] $. 
Next, we  use the a bidirectional LSTM to capture context-specific information for each word,
resulting in the word-level features:
\begin{center}
\vspace{-3mm}
\small
\begin{align}
{\bf{h}}_i^w = {[{{\mathop{\rm biLSTM}\nolimits} _1}({\bf{v}}_0,...,{\bf{v}}_n)]_i}
\end{align}
\vspace{-3mm}
\end{center}

Such representations are then used as inputs to a second LSTM to generate span-level features:
Inspired by \cite{kong2015segmental}, we compute all possible span embeddings efficiently with time complexity $\mathcal{O}(cn)$ using dynamic programming, 
with $n$ being the number of words in the input $\boldsymbol{x}$ and $c$ being the maximal length of a mention.
%
\begin{center}
\vspace{-3mm}
\small
\begin{align}
{\bf{h}}_{i:j}^s = {{\mathop{\rm biLSTM}\nolimits} _2}({\bf{h}}_i^w,...,{\bf{h}}_j^w)
\end{align}
\vspace{-3mm}
\end{center}




Recall that there are 4 types of hyperedges in our hypergraph, over which we can define the score functions.
Since every valid mention hyperpath contains the first and second kind of hyperedges, defining scores over such hyperedges are unnecessary
as their scores would serve as a constant factor that can be eliminated in the overall loss function of the log-linear model.
Thus we only need to define the score functions on the latter two types of hyperedges.
For hyperedges that only involve two nodes, we use a linear layer to compute their scores:
\vspace{-3mm}
\begin{center}
\small
\begin{align}
\phi (\{\boldsymbol{\mathsf{T}}^k_i\rightarrow\boldsymbol{\mathsf{X}}\}, \boldsymbol{x}) = {\mathbf{W}^{(k)}_{{\scaleto{\rm{TX}}{3.5pt}}}} \cdot {\bf{h}}_i^w
\\
\phi (\{\boldsymbol{\mathsf{T}}^k_i\rightarrow\boldsymbol{\mathsf{I}}^k_{i,j}\},\boldsymbol{x}) =  {\mathbf{W}^{(k)}_{{\scaleto{\rm{TI}}{3.5pt}}}} \cdot {\bf{h}}_i^w
\\
\phi (\{\boldsymbol{\mathsf{I}}^k_{i,j}\rightarrow\boldsymbol{\mathsf{I}}^k_{i,j+1}\},\boldsymbol{x}) =  {\mathbf{W}^{(k)}_{{\scaleto{\rm{II}}{3.5pt}}}} \!\!\!\cdot\! [{\bf{h}}_{i:j}^s,{\bf{h}}_{i:j+1}^s]
\label{eq:II}
\\
\phi (\{\boldsymbol{\mathsf{I}}^k_{i,j}\rightarrow\boldsymbol{\mathsf{X}}\},\boldsymbol{x}) =  {\mathbf{W}^{(k)}_{{\scaleto{\rm{IX}}{3.5pt}}}} \cdot {\bf{h}}_{i:j}^s 
\end{align}
\end{center}
where matrices 
$ \mathbf{W}_{{\scaleto{\rm{TX}}{3.5pt}}}, \mathbf{W}_{{\scaleto{\rm{TI}}{3.5pt}}} \in \mathbb{R}^{{d_1} \times m} $,
$ \mathbf{W}_{{\scaleto{\rm{II}}{3.5pt}}} \in \mathbb{R}^{{2d_2} \times m} $,
$ \mathbf{W}_{{\scaleto{\rm{IX}}{3.5pt}}} \in \mathbb{R}^{{d_2} \times m} $, with superscript $(k)$ referring to the $k$-th column of the matrix,
$d_1$ is the dimension of $\mathbf{h}^w$, $d_2$ is the dimension of $\mathbf{h}^s$, and $m$ is the number of mention types.

For  the hyperedges that involve more than two nodes, the score is computed as follows:
\begin{center}
\vspace{-3mm}
\small
\begin{align}
&\phi (\{\boldsymbol{\mathsf{I}}^k_{i,j}\rightarrow(\boldsymbol{\mathsf{X}},\boldsymbol{\mathsf{I}}^k_{i,j+1})\}, \boldsymbol{x})
\nonumber
\\
= &\ \  {\mathbf{W}'^{(k)}_{{\scaleto{\rm{II}}{3.5pt}}}} \cdot [{\bf{h}}_{i:j}^s,{\bf{h}}_{i:j+1}^s]  + {\mathbf{W}'^{(k)}_{{\scaleto{\rm{IX}}{3.5pt}}}} \cdot {\bf{h}}_{i:j}^s
\label{eq:IIX}
\end{align}
\end{center}
where
$ \mathbf{W'}_{{\scaleto{\rm{II}}{3.5pt}}} \in \mathbb{R}^{{2d_2} \times m} $,
$ \mathbf{W'}_{{\scaleto{\rm{IX}}{3.5pt}}} \in \mathbb{R}^{{d_2} \times m} $.
Note that in this work, we set $\mathbf{W'}_{{\scaleto{\rm{II}}{3.5pt}}} = \mathbf{W}_{{\scaleto{\rm{II}}{3.5pt}}} $ and $\mathbf{W'}_{{\scaleto{\rm{IX}}{3.5pt}}} = \mathbf{W}_{{\scaleto{\rm{IX}}{3.5pt}}} $ to reduce the number of free parameters.

Learning uses stochastic gradient descent with the update rule of Adam \cite{kingma2014adam} and a gradient clipping of 3.0.
Dropout \cite{srivastava2014dropout} for input vectors $\mathbf{v}$ and $\ell_2$ regularization are used to reduce overfitting; both are tuned during the development process.

\subsection{Character-level Representation}

To make fair comparisons with recent models \cite{N18-1131,nest-bw-em18} that additionally incorporate character-level components in capturing orthographic and morphological features of words, we follow \citet{lample2016neural} to use a bidirectional LSTM that takes the  character embeddings as input.
Specifically, the character-level representation $\mathbf{ch}_i$ for each word is obtained by concatenating the last hidden vectors of the forward and backward LSTMs.
When this component is activated, the representation of each word is changed to: 
$\mathbf{v}_i= [\mathbf{e}_i, \mathbf{p}_i, \mathbf{ch}_i] $.

\section{Inference}

Inference can be done efficiently using a generalized inside-outside style message-passing algorithm \cite{baker1979trainable}.
The partition function of (\ref{eq:loglinear}) can be computed using the inside algorithm applied to the complete hypergraph $\mathcal{G}$, where we traverse from leaf nodes $\boldsymbol{\mathsf{X}}$ to the root node $\boldsymbol{\mathsf{A}}_1$, passing messages to a parent node $\boldsymbol{\mathsf{p}}$ from its child nodes:
\begin{center}
\small
\begin{align}
\mu[\boldsymbol{\mathsf{p}}]
\!
\!
\leftarrow
\!
\log
\Big(\!\!\!\!\!
\sum_{e: h(e)\equiv \boldsymbol{\mathsf{p}}}
\!\!\!\!
\exp 
\big(
\psi(e,\boldsymbol{x})
+
\!\!\!\!
\sum_{\boldsymbol{\mathsf{c}}\in \mathcal{T}(e)} 
\!\!\!\!
\mu[\boldsymbol{\mathsf{c}}]
\big)
\Big)
\label{eqn:mp}
\end{align}
\end{center}
where $h(e)$ is the head of the hyperedge $e$, and $\mathcal{T}(e)$ is the collection of nodes that form the tail of $e$ -- they are the child nodes of $h(e)$ given $e$.
The message passing step for the outside algorithm can be defined analogously.
It can be verified that such a message passing algorithm, that is analogous to the sum-product belief propagation algorithm \cite{kschischang2001factor} used in {\color{black} standard graphical models}, will converge after one forward and one backward pass. 

For decoding, we perform the standard MAP inference on top of the complete hypergraph to find the most probable hyperpath.
The resulting procedure is similar to the max-product message passing algorithm, where we consider only the feature function $\phi$ for constructing the messages:
\begin{center}
\small
\begin{align}
\mu[\boldsymbol{\mathsf{p}}]
\leftarrow
\max_{e: h(e)\equiv \boldsymbol{\mathsf{p}}} 
\Big(
\phi(e,\boldsymbol{x})
+
\sum_{\boldsymbol{\mathsf{c}}\in \mathcal{T}(e)} 
\mu[\boldsymbol{\mathsf{c}}]
\Big)
\end{align}
\end{center}

During inference, each node corresponds to a sum/max computation.
Since one node is incident to 3 hyperedges maximally, the time complexity of inference algorithm can be implied by the number of nodes in the graph, 
which is $\mathcal{O}(cmn)$, where $c$ is the maximal length for any mention. 
This complexity is the same as that of a zero-th order semi-Markov CRF model \cite{sarawagi2005semi}.
Please refer to the supplementary material for a detailed explanation of the inference algorithm.

\begin{table}[t]
\centering
\scalebox{0.7}{
\begin{tabular}{r|rr|rr}
\multicolumn{1}{c|}{}
   & \multicolumn{2}{c|}{ACE-2004} & \multicolumn{2}{c}{GENIA}   \\  
\multicolumn{1}{c|}{}
             & \multicolumn{1}{c}{Train (\%)}& \multicolumn{1}{c|}{Test (\%)}  & \multicolumn{1}{c}{Train (\%)} & \multicolumn{1}{c}{Test (\%)}     \\ 
\hline

\# sentences  & 6,799 {\color{white} (00)} &879{\color{white} (00)} & 14,836 {\color{white} (00)}   & 1,855 {\color{white} (00)}  \\ 

\textit{with o.l. } & 2,683 (39)  & 272 (42)  & 3,199 (22)  & 448 (24) \\

\hline
\# mentions & 22,207 {\color{white} (00)}  & 3,031 {\color{white} (00)} & 46,473 {\color{white} (00)}  & 5,600 {\color{white} (00)} \\

\textit{o.l.} & 10,170 (46) & 1,418 (47) & 8,337 (18)  & 1,217 (22) \\

\textit{o.l. (st)} & 5,431 (24)  & 780 (26)  & 4,613 (10)  & 634 (11)  \\

\textit{o.l. (st \& slb)} & 2,188 (10)  & 307 (10)  & 2,133 ({\color{white} 0}5)  & 287 ({\color{white} 0}5)  \\

\textit{lengh $> 6$ } & 1,439 ({\color{white} 0}6)  & 199 ({\color{white} 0}7)  & 2,449 ({\color{white} 0}5)  & 301 ({\color{white} 0}5) \\

\hspace{2mm} \textit{max lengh }  & 57 {\color{white} (00)} & 43 {\color{white} (00)} & 28 {\color{white} (00)} & 19 {\color{white} (00)}  \\
\end{tabular}
}
\caption{Statistics (ACE04, GENIA). \textit{o.l.}: overlapping mentions, {\em st/slb}: same type/left boundary.}
\label{tab:stat}
\end{table}

\section{Experiments}



\subsection{Datasets}


We mainly evaluate our models on the standard ACE-2004, ACE-2005 \cite{doddington2004automatic}, and GENIA \cite{kim2003genia} datasets  with the same splits used by previous works \cite{lu2015joint,muis2017labeling}.
Sample data statistics of these  datasets are listed in Table \ref{tab:stat}. 
\footnote{See supplementary material for complete data statistics.}
We can see that overlapping mentions frequently appear in such datasets. 
{\color{black} 
For ACE2004, over 46\% of the mentions overlap with one another.
GENIA focuses on biomedical entity recognition\footnote{Following previous works, we used version 3.02p which comes with annotated POS tags \cite{tateisi2004part} .
Following  \cite{finkel2009nested}, we collapse \textit{DNA}, \textit{RNA} and \textit{protein} subtypes into \textit{DNA}, \textit{RNA} and \textit{protein} respectively, keep cell line and cell type and remove  mentions of other types.}
and overlapping mentions are also common in it.
Most mentions (over 93\%) are not longer than 6 tokens which we select as  maximal length ($c$) for the restricted models.}

\begin{table*}[t]
\centering
\scalebox{0.66}{
\begin{tabular}{c|l|ccc|ccc|ccc}
\multicolumn{2}{c|}{}& \multicolumn{3}{c|}{ACE-2004} & \multicolumn{3}{c|}{ACE-2005}   & \multicolumn{3}{c}{GENIA} \\
\multicolumn{2}{c|}{}&$P$ & $R$ & $F_1$  & $P$ & $R$ & $F_1$ & $P$ & $R$ & $F_1$ \\ 
\hline
\multirow{9}{*}{Non-Neural}&CRF (\textsc{linear})       &     71.8  &  40.8 & 52.1    & 69.5 & 44.5 & 54.2  & 77.1 & 63.3 & 69.5  \\
&CRF (\textsc{cascaded})   &   78.4   &  46.4 & 58.3 &   74.8 & 49.1 & 59.3  & 75.9 & 66.1 & 70.6  \\
&Semi-CRF ($c$=$6$)  & 76.1  &  41.4 & 53.6    & 72.8 & 45.0 & 55.6 &  74.5 &  66.0 & 70.0  \\
&Semi-CRF ($c$=$n$)     & 66.7  &  42.0 & 51.5   & 67.5 & 46.1 & 54.8  &  74.2  &  65.8 & 69.7  \\
&\citet{finkel2009nested}  & - & - & - & -& -& -  & 75.4   &  65.9 & 70.3 \\
&\citet{lu2015joint}    &  70.0  &  56.9 & 62.8 & 66.3 & 59.2 & 62.5 & 74.2 & 66.7 & 70.3  \\
&\citet{muis2017labeling}  & 72.7   &  58.0 & 64.5     & 69.1 &  58.1 & 63.1 & 75.4 & 66.8 & 70.8  \\
\cdashline{2-11}
&SH (-\textsc{nn}, $c$=$6$)    & 69.4 &  57.0 & 62.0  & 70.3 &  55.8 & 62.2 & 77.0 & 66.1 & 71.1   \\
&SH (-\textsc{nn}, $c$=$n$)   & 71.1 &  60.6 & 65.4  & 69.5 &  60.7 & 64.8  & 76.2 & 67.5  & 71.6 \\
 \hline
 \hline
\multirow{4}{*}{Neural}&FOFE \cite{xu2017local} ($c$=$6$)   & 68.2 &  54.3 & 60.5   & 67.4 &  55.1 & 60.6  & 71.2 & 64.3 & 67.6 \\
&FOFE \cite{xu2017local} ($c$=$n$)    & 57.3 &  46.8 & 51.5  & 56.3 &  44.6 & 49.8 & 63.2 & 59.3 & 61.2 \\
& \citet{N18-1079} & 73.6 & 71.8 & 72.7 & 70.6 & 70.4 & 70.5 & \bf{79.8} & 68.2 & 73.6 \\
& \citet{N18-1131} 
\footnotemark 
& - & - & - & 74.2 & 70.3 & 72.2 & 78.5 & 71.3 & 74.7 \\
& \citet{nest-bw-em18} & 74.9 & 71.8 & 73.3 & 74.5 & 71.5 & 73.0 & 78.0 & 70.2  & 73.9 \\
\cdashline{2-11}
&SH ($c$=$6$)    & 79.1 &  67.3 & 72.7   & 75.7 &  69.6 & 72.5  & 76.6 & 71.0 & 73.7 \\
&SH ($c$=$6$) + \textit{char}  & \bf{80.1} &  67.5 & 73.3   & 75.9 &  70.0 & 72.8  & 76.8 & 71.8 & 74.2 \\
&SH ($c$=$n$)   & 77.7  & 72.1 & 74.5  & 76.6 &  71.9 & 74.2   & 76.1 &  72.9 &74.5 \\
&SH ($c$=$n$)  + \textit{char} & 78.0  & \bf{72.4} & \bf{75.1}  & \bf{76.8} &  \bf{72.3} &\bf{74.5}   & 77.0 &  \bf{73.3} & \bf{75.1}  \\
\end{tabular}
}
\caption{Main results. SH: segmental hypergraphs (our approach). }
\label{tab:result}
\end{table*}

\subsection{Baseline Approaches}

We consider the following baseline models:

\squishlist

\item CRF (\textsc{linear}): a linear-chain CRF model. 
Since the linear-chain CRF cannot handle overlapping structures, we only use the outer-most mentions for learning.
Specifically, every outer-most mention is labeled based on the BILOU tagging scheme, which was empirically shown to be better than the BIO scheme \cite{ratinov2009design}.

\item CRF (\textsc{cascaded}): the cascaded CRF based approach following \cite{alex2007recognising}. 
Note that this approach cannot model the overlapping mentions of the same type.

\item Semi-CRF: the semi-Markov CRF model \cite{sarawagi2005semi}.
The semi-CRF model is also only trained on the outer-most mentions.
It can also capture span-level features defined over a complete segment.
Similar to our model, semi-CRF typically comes with a length restriction ($c$) which indicates the maximal length of a mention.

\item \citet{finkel2009nested}: a parsing-based approach for recognizing nested mentions that reported results on the GENIA dataset.

\item \citet{lu2015joint}: the model that makes use of mention hypergraphs for recognizing overlapping mentions.

\item \citet{muis2017labeling}: the model that makes use of mention separators to tag gaps between words for recognizing overlapping mentions.

\item FOFE \cite{xu2017local}: a local classifier based on neural networks that runs on every possible span to detect mentions.
The maximal mention length ($c$) can also be used here.

\footnotetext{Note that in ACE2005, \citet{N18-1131} did their experiments with a different split than \citet{lu2015joint} which we follow as our split.}

\item \citet{N18-1079}: a hypergraph-based model that uses LSTM for learning feature representations.

\item \citet{N18-1131}: a cascaded model that makes use of multiple LSTM-CRF layers to recognize mentions in an inside-out manner. 

\item \citet{nest-bw-em18}: a neural transition-based model that construct nested mentions through a sequence of actions.

\item SH (-\textsc{nn}): a non-neural version of our segmental hypergraph model that excludes the LSTMs but employs handcrafted features.
\footnote{To make a proper comparison, we use the same handcrafted features used by \cite{lu2015joint}, which were proven effective in previous approaches. 
}

\squishend

As discussed earlier, we also evaluate the variants of our model that takes character-level representations (+$\textit{char}$).

\subsection{\color{black}{Training}}

Pre-trained embeddings GloVe \cite{pennington2014glove} of dimension 100 are used to initialize the trainable word vectors for experiments in ACE and GENIA datasets.\footnote{We also additionally tried using embeddings trained on PubMed for GENIA but the performance was comparable.}
The embeddings for POS tags are initialized randomly with dimension 32. 
Early stopping is used based on the performance of development set.
The value $\beta$ used in softmax-margin is chosen from [1, 3]  with step size 0.5.

\subsection{Experimental Results}

\begin{table}[t!]
  \centering
  \scalebox{0.68}{
\begin{tabular}{cl|cc|cc|cc}
\multicolumn{2}{c}{}   & \multicolumn{2}{|c|}{ACE-2004} & \multicolumn{2}{c|}{ACE-2005} & \multicolumn{2}{c}{GENIA}  \\ 
&& ($c$=$6$) &  ($c$=$n$) &  ($c$=$6$) &  ($c$=$n$) &  ($c$=$6$) &  ($c$=$n$) \\
\hline
\multicolumn{2}{c|}{SH} & 72.7 & 74.5 & 72.5 & 74.2 & 73.7 & 74.5 \\
&-\textsc{d}& 71.5 & 73.1 & 71.3 & 72.9 & 72.1 & 72.8 \\
&-\textsc{sm} & 72.0 & 73.3  & 71.8  & 73.5 & 72.4  & 73.3 \\
&-\textsc{p} & 71.5 & 72.7 & 71.2  & 73.0 & 72.0 & 73.2  
\end{tabular}
}
\caption{Results of various ablations. \textsc{d}: dropout, \textsc{sm}: softmax-margin, \textsc{p}: pre-trained embeddings.}
\label{tab:ablation}
\end{table}

Main results can be found in Table \ref{tab:result}.
Using the same set of handcrafted features, our unrestricted non-neural model SH (-\textsc{nn}, $c$=$n$) achieves the best performance compared with other non-neural models, 
revealing the effectiveness of our newly proposed segmental hypergraph representation.
It achieves around 1-2\% gain in terms of $F_1$ compared with mention hypergraph of \citet{lu2015joint} and mention separator of \citet{muis2017labeling}, 
showing the necessity of eliminating structural ambiguity.
CRF (\textsc{linear}) and Semi-CRF do not perform well due to incapability of handling overlapping mentions. 
In contrast, the pipeline approach CRF (\textsc{cascaded}) performs better.

Our unrestricted neural segmental hypergraph model SH ($c$=$n$) already achieves the best results among 
all previous models in ACE datasets, showing the effectiveness of our neural segmental hypergraph.
The improvement mainly comes from its ability to recall more mentions.
In GENIA, even without using external features like Brown clustering features as all non-neural models do, our neural models still get significant improvements. 
Compared with the non-neural SH (-\textsc{nn}) which has around 4.2M parameters, our neural model SH only has  1.9M parameters yet it still performs better.
We empirically see that the representations learned by LSTM can better capture complex contextual dependencies in sentences.
The character-level representations (+ $\textit{char}$) make both restricted and unrestricted SH perform even better.
Particularly, SH ($c$=$n$) + $\textit{char}$ achieves the best results in all datasets 
compared with other recent neural models  \cite{N18-1079,N18-1131,nest-bw-em18}.

One hypothesis we may have is that, without length restriction, a model will enjoy the benefit of recalling more long mentions, but also will be exposed to more false positives.
This poses a challenge for a model -- whether it is capable of balancing these two factors. 
Empirically, we find that the length restriction ($c$=$6$) improves the precision of semi-CRF and SH at the expense of the recall, providing some evidence to support the hypothesis. 
However, in terms of $F_1$, the unrestricted semi-CRF performs worse while unrestricted SH performs better compared to their restricted counterparts. 
The reason is that the span-level handcrafted features that the semi-CRF relies on can be very sparse when mentions are overly long.
We empirically found this issue is alleviated in the model SH (-\textsc{nn}), possibly due to its ability in capturing interactions between neighboring spans.
Even with length restriction, SH still yields competitive results, making it attractive in processing large-scale datasets considering its linear time complexity.
{\color{black}Furthermore, we find that as $c$ increases, SH performs better consistently in terms of $F_1$. 
The choice of $c$ then becomes a tradeoff between time complexity and performance.
Please refer to the supplementary material for details.
}

Compared with the local approach FOFE, our global approach gives a much better performance, showing its effectiveness in capturing interactions between spans.
Moreover, FOFE's performance suffers significantly in the absence of the length restriction. The reason is that it will generate much more negative training instances under this setting, which makes its learning more challenging.

\begin{table}[t!]
\centering
\scalebox{0.68}
{
%
\begin{tabular}{l|ccc|ccc|c}
    & \multicolumn{3}{c|}{{Overlapping}} & \multicolumn{3}{c|}{{Non-Overlapping}} &\multirow{2}{*}{{\em w/s}}    \\ 
             &$P$ & $R$ & $F_1$ & $P$ & $R$ & $F_1$  &   \\ 
 \hline
\citet{lu2015joint}         &    68.1   &  52.6 & 59.4    & 64.1  &  65.1 & 64.6 & 503   \\
\citet{muis2017labeling}          & 70.4  &  55.0 & 61.8    & 67.2   &  63.4 & 65.2  & 253     \\
\citet{nest-bw-em18} & 77.4 & 70.5 &  73.8 & 76.1 & 69.6 &  72.7 & 1445 \\ 
\hline

 SH (c=$6$)  & 80.2 & 68.3 & 73.8 & 74.8 & 70.0 & 72.3    &  248  \\
 SH (c=$n$)  & 80.6 & 73.6 & 76.9 & 75.5 & 71.5 & 73.4   &  157    \\
 
\end{tabular}
}
\caption{Results on different types of sentences (ACE05), {\em w/s}: \# of words decoded per second.}
\label{tab:overlap}
\end{table}

\subsection{Additional Analyses}

To understand our model better, we conduct some further experiments in this section. 

\subsubsection*{Ablation study}

We first conduct an ablation study by removing dropout, softmax-margin and pre-trained embeddings from our model respectively.
The results are shown in Table \ref{tab:ablation}.
The dropout and pre-trained embeddings can improve the {\color{black}performance} of our model significantly and this behavior is consistent with previous neural models for NER \cite{chiu2016named,lample2016neural}.
Meanwhile, our new cost function based on softmax margin training also contributes significantly to the good performance of our model across these datasets.

\subsubsection*{How well does it handle overlapping mentions?}

To further understand how well our model can handle overlapping mentions, we 
split the test data into two portions:
sentences with and without overlapping mentions.
We compare our model with the two state-of-the-art models and report results on ACE-05 in Table \ref{tab:overlap}.\footnote{Full results are listed in the supplementary material.}
In both portions, SH achieves significant improvements, especially in the portion with overlapping mentions.
This observation indicates that our model can better capture the structure of overlapping mentions than these two previous models.
{\color{black}It also helps explain why the margin of improvement is larger in ACE than in GENIA since the former has more overlapping mentions than the latter, as shown in Table \ref{tab:stat}.}
Compared with the model with length restriction $c$, the unrestricted model mainly benefits from its ability to recall more overlapping mentions.

\subsubsection*{Running time}

Since other compared models also feature linear time complexity (see Table \ref{tab:model_comparison}),
we examine the decoding speed in terms of {\color{black} the number of words} processed per second.
We re-implement the models of \citet{lu2015joint} and \citet{muis2017labeling} using the same platform as ours (PyTorch) and run them on the same machine (CPU: Intel i5 2.7 GHz).
The model of \cite{nest-bw-em18} is also tested with the same environment. 
Results on ACE-05 are listed in Table \ref{tab:overlap}.
The length bound ($c$=6) makes our model much faster, resulting in a  speed comparable to the model of \citet{muis2017labeling}.
The transition-based model by \cite{nest-bw-em18} has the best scalability partially because of its greedy strategy for decoding.

\subsubsection*{What if the data has no overlapping mentions?}

\begin{table}[t!]
\centering
\scalebox{0.9}
{
\begin{tabular}{l|c}
Model & $F_1$ \\
  \hline
  
SH ($c$=$6$)   & 89.6      \\
SH ($c$=$6$) + \textit{char}   & 90.5      \\
SH ($c$=$n$)   & 89.2       \\
SH ($c$=$n$)  + \textit{char}  & 90.2     \\
\hdashline
\citet{collobert2011natural} & 88.7 \\
\citet{chiu2016named} & 90.9 \\
\citet{lample2016neural}& 90.9\\
\citet{ma-hovy:2016:P16-1} & 91.2 \\
\citet{xu2017local} & 90.7 \\
\citet{strubell2017fast} & 90.5
\end{tabular}
}
\caption{Additional results on CoNLL-2003.}
\label{tab:ner-conll}
\end{table}

To assess the robustness of our model and understand whether it could serve as a general mention extraction model,
we additionally evaluate our model on CoNLL 2003  dataset which is annotated with non-overlapping mentions only.
We compared our model with recent state-of-the-art neural network based models.
For a fair comparison, we used the \newcite{collobert2011natural} embeddings widely used by previous models, and ignored POS tag features even though they are available.
Results are  in Table \ref{tab:ner-conll}. Only neural models without using external features are included. 
\footnote{See the supplementary material for complete results.}
By only relying on word (and character) embeddings, our model achieves competitive results compared with other state-of-the-art neural models that also do not exploit external features, 
yet these models are mostly designed to handle only non-overlapping mentions.
The only exception is the FOFE approach by \cite{xu2017local} as we discussed earlier.

\subsubsection*{Notes on mention interactions}

The dependencies between overlapping mentions can be very beneficial.
SH can capture a specific kind of interaction between neighboring spans.
Such interactions happen between mentions that share the same type and the same left boundary. 
As we can see from the sentence in Figure \ref{fig:example}, one mention could also serve as a pre-modifier for another mention and both could share the same type.
As shown in Table \ref{tab:stat}, there are over 8\% such mentions in ACE  and over 4\% in GENIA. 
Specifically, {\color{black} SH relies on the hyperedges between $\boldsymbol{\mathsf{I}}$ nodes to capture such interactions explicitly.}
To verify the effectiveness of {\color{black}this connection}, we zero the weights between $\boldsymbol{\mathsf{I}}$ nodes.
The ablated model only achieves around 70.0\% in ACEs and 71.4\% in GENIA, implying the impact of this {\color{black}dependency connection}.
On the other hand, it also reveals the potential direction of improving SH by explicitly modeling more dependencies between mentions, such as the dependencies between mentions with different types.
LSTM that serves as feature representation may capture such interactions implicitly, but building the connections could still be an important aspect for improvement.


\section{Conclusion and Future Work}

In this work, we propose a novel neural segmental hypergraph model that is able to capture overlapping mentions.
We show that our model has some theoretical advantages over previous state-of-the-art approaches for recognizing overlapping mentions.
Through extensive experiments,  we show that our model is general and robust in handling both overlapping and non-overlapping mentions. 
The model achieves the state-of-the-art results in three standard datasets for recognizing overlapping mentions.
We anticipate this model could be leveraged in other similar sequence modeling tasks that involve predicting overlapping structures such as recognizing overlapping and discontinuous entities \cite{muis2016learning} which frequently exist in the biomedical domain.

\section*{Acknowledgements}

We thank the anonymous reviewers for their valuable comments. 
This work was done after the first author visited Singapore University of Technology and Design.
This work is supported by Singapore Ministry of Education Academic Research Fund (AcRF) Tier 2 Project MOE2017-T2-1-156.

\bibliography{semi}
\bibliographystyle{acl_natbib_nourl}

\end{document}